\documentclass[runningheads]{llncs}
\usepackage{marvosym}

\usepackage{amsmath}
\usepackage{graphicx}
\usepackage{multirow}
\newcommand{\tabincell}[2]{\begin{tabular}{@{}#1@{}}#2\end{tabular}}
\begin{document}
\title{Deep Fiber Clustering: Anatomically Informed Unsupervised Deep Learning for Fast and Effective White Matter Parcellation\thanks{We acknowledge funding provided by the following National Institutes of Health (NIH) grants: R01MH125860, R01MH119222, R01MH074794, and P41EB015902.}}
\titlerunning{Deep Fiber Clustering}
%
\author{Yuqian Chen\inst{1,2} \and
Chaoyi Zhang\inst{2} \and
Yang Song\inst{3} \and
Nikos Makris\inst{1} \and
Yogesh Rathi\inst{1} \and
Weidong Cai\inst{2} \and
Fan Zhang\inst{1}$^($\textsuperscript{\Letter}$^)$ \and
Lauren J. O’Donnell\inst{1}
}

\authorrunning{Y. Chen et al.}
%

%
\institute{Harvard Medical School, MA, USA \\ \email{fzhang@bwh.harvard.edu} \and The University of Sydney, NSW, Australia  \and The University of New South Wales, NSW, Australia}
\maketitle              
\begin{abstract}
White matter fiber clustering (WMFC) enables parcellation of white matter tractography for applications such as disease classification and anatomical tract segmentation. However, the lack of ground truth and the ambiguity of fiber data (the points along a fiber can equivalently be represented in forward or reverse order) pose challenges to this task. We propose a novel WMFC framework based on unsupervised deep learning. We solve the unsupervised clustering problem as a self-supervised learning task. Specifically, we use a convolutional neural network to learn embeddings of input fibers, using pairwise fiber distances as pseudo annotations. This enables WMFC that is insensitive to fiber point ordering. In addition, anatomical coherence of fiber clusters is improved by incorporating brain anatomical segmentation data. The proposed framework enables outlier removal in a natural way by rejecting fibers with low cluster assignment probability. We train and evaluate our method using 200 datasets from the Human Connectome Project. Results demonstrate superior performance and efficiency of the proposed approach.

\keywords{Diffusion MRI  \and Tractography \and Fiber clustering \and Deep embedding \and Self-supervised learning.}
\end{abstract}
\section{Introduction}
Diffusion magnetic resonance imaging (dMRI) \cite{C1} uniquely enables mapping of the brain's white matter fiber tracts via tractography \cite{C2}, to study the brain's connections in health and disease \cite{C3}. Tractography of a single brain can generate hundreds of thousands of streamlines (fibers), which are not immediately useful to clinicians or researchers. Therefore, tractography parcellation, i.e. dividing the massive number of tractography fibers into multiple subdivisions, is needed.

One widely used tractography parcellation strategy, white matter fiber clustering (WMFC), groups fiber streamlines with similar geometric trajectory into clusters \cite{C4}. WMFC is useful in applications such as disease classification \cite{zhang2018whole}, anatomical tract identification \cite{C31} and neurosurgical brain mapping \cite{tuncc2016individualized}. In general, WMFC first computes pairwise fiber geometric similarities, then applies a computational clustering method to group similar fibers into clusters \cite{C5,vazquez2020ffclust,C7}. Existing WMFC methods show good performance, but key challenges remain. First, it is computationally expensive to compute pairwise fiber geometric similarities. Second, the computation of fiber similarity is sensitive to the order of points along the fibers, even though a fiber can equivalently start from either end \cite{C5}. Third, false positive fibers are prevalent in tractography; thus outlier fiber removal is needed to filter undesired fibers from the clustering result \cite{legarreta2020tractography,C6}. Fourth, it is a challenge for WMFC to use all available information to improve cluster anatomical quality: most methods use either fiber spatial coordinate information \cite{C5,C7} or anatomical information about brain regions that fibers pass through \cite{C8}. Fifth, WMFC methods should ideally consider inter-subject correspondence of fiber clusters, which is essential for group-wise analysis \cite{C9}. To achieve this goal, some studies perform WMFC across subjects (to form an atlas) and predict clusters of new subjects with correspondence to the atlas \cite{C11,C12,C10}, while other approaches first perform within-subject WMFC then match (or cluster) the fiber clusters across subjects \cite{C5,guevara2012automatic,huerta2020inter,C8}.

In computer vision, clustering has been extensively studied as an unsupervised learning task \cite{C17,C15,C13,C14,C16}, which requires a data feature representation and similarity computation between the features for cluster assignment. Autoencoder-based approaches are popularly used for unsupervised clustering \cite{C15,C13,C14}. The Deep Embedding Clustering (DEC) framework  performs simultaneous embedding of input data and cluster assignments in an end-to-end way \cite{C14}. Deep Convolutional Embedded Clustering (DCEC) is an extension of DEC to the image clustering task \cite{C15}. In addition to autoencoder approaches, \cite{C17} and \cite{C16} also realized joint embedding learning and cluster assignments by alternative feature learning and traditional clustering, which is time consuming.

Self-supervised learning is a promising subclass of unsupervised learning that shows advanced performance in many applications \cite{C18,C21}. It aims to learn high-level features without requiring manual annotations. This is achieved by designing pretext tasks, such as predicting context \cite{C19} or image rotation \cite{C20}, and giving the network pseudo annotations generated from the input itself. The high-level representations learned from the pretext task can then be transferred to downstream tasks such as clustering. Therefore, besides the classical autoencoder network, the self-supervised learning framework can also be a promising approach to learn deep embeddings of inputs.

Considering the advances of deep neural networks in feature extraction, deep learning is a promising direction for WMFC. In related work, multiple deep learning methods have been proposed for white matter tractography segmentation \cite{C26,wasserthal2018tractseg,xu2020vector,C22}. In \cite{C26,wasserthal2018tractseg,C22}, known fiber labels are provided for training. One proposed method \cite{xu2020vector} has shown the potential of unsupervised deep learning for fiber clustering; however, the anatomical utility of this approach was not tested as results were limited to a maximum of 11 clusters in the whole brain. The goal of our study is to propose an anatomically meaningful unsupervised deep learning framework, Deep Fiber Clustering (DFC), for fast and effective white matter fiber clustering. The paper has four contributions. First, we propose a novel deep learning pipeline that adopts self-supervised learning for deep embedding and achieves joint representation learning and cluster assignment. Second, anatomical information is incorporated into the neural network to improve cluster anatomical coherence. Third, outliers are removed by rejecting fibers with low soft label assignment probabilities. Our approach automatically creates a multi-subject fiber cluster atlas that is applied for white matter parcellation of new subjects. Finally, our approach has demonstrated superior performance and efficiency via evaluations on a large scale dataset.

\begin{figure}[!t]\centering 
\includegraphics[width=9.5cm]{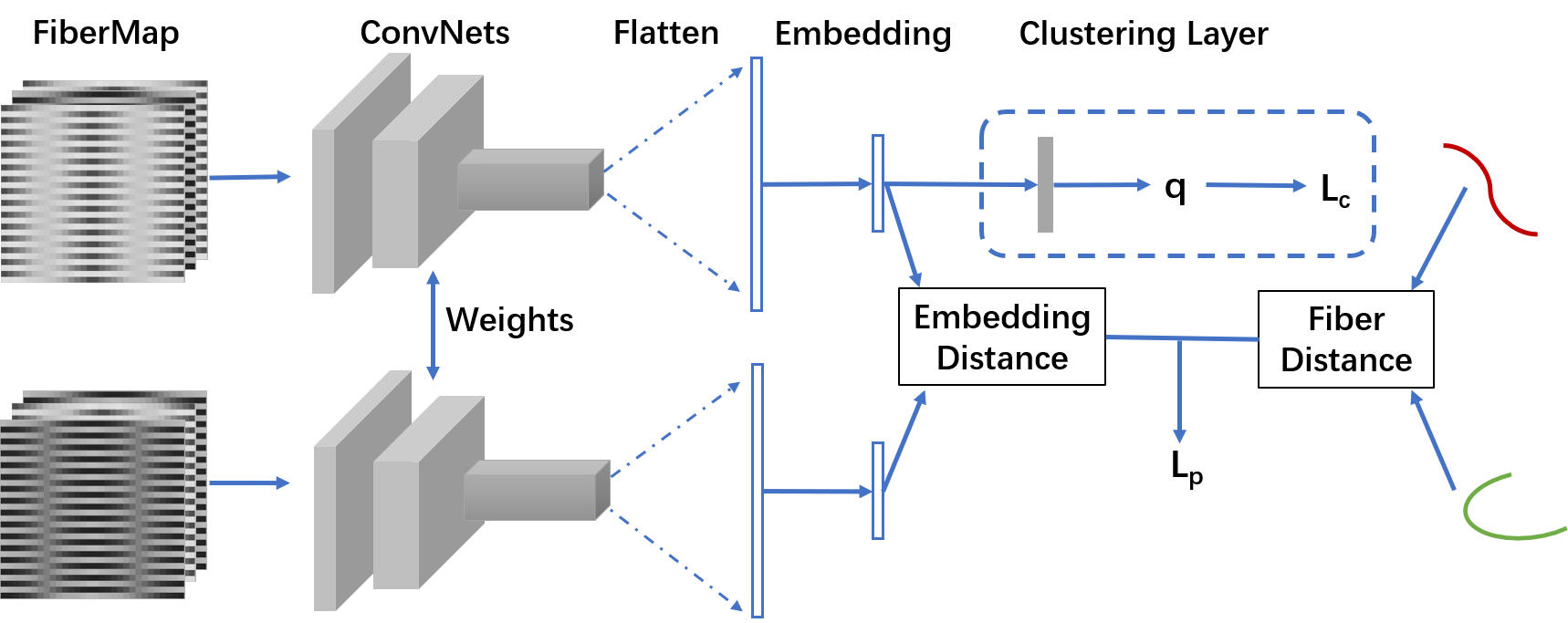}
\caption{Overview of our DFC framework. A self-supervised learning strategy is adopted with the pretext task of pairwise fiber distance prediction. In the pretraining stage, a pair of FiberMaps are encoded as embeddings with Siamese Networks, and prediction loss ($L_p$) is calculated based on the difference between embedding distance and fiber distance. In the clustering stage, a clustering layer is connected to the embedding layer and generates soft label assignment (as shown in the dashed box). A KL divergence loss ($L_c$) and the prediction loss are combined to optimize the neural network.} \label{fig1}
\end{figure}

\section{Methods}
As shown in Fig. \ref{fig1}, our training pipeline includes two stages, \textit{pretraining} and \textit{clustering}. In the pretraining stage (Sec. 2.1), a CNN is trained in a self-supervised way with a designed pretext task to obtain deep embeddings. After that, $\mathnormal{k}$-means clustering is performed on the embeddings to get initial clusters, which is performed only once during training. In the clustering stage (Sec. 2.2), the clustering results are fine-tuned in a self-learning manner and cluster centroids are automatically optimized as parameters of the network. During network inference, when the model is applied to a new subject, cluster assignments are obtained from the network directly in an end-to-end way without any $\mathnormal{k}$-means clustering.

In this work, we adopt the FiberMap fiber representation \cite{C22}, which was found to be effective for tractography segmentation in supervised learning. One benefit of using FiberMap is that it is a 2D multi-channel feature descriptor (analogous to a RGB image); thus it can be effectively processed by CNNs.

\subsection{Self-supervised Deep Embedding}
We propose a novel self-supervised learning strategy for learning deep fiber embeddings. The goal is to obtain embeddings with similar distances to fiber distances in the brain space, enabling subsequent WMFC in the embedding space. (We note that a DCEC model with a convolutional autoencoder could be adopted here for unsupervised WMFC, but as we show in the Results, this straightforward approach is sensitive to fiber point ordering.) To learn the embeddings, a pretext task is first designed to predict the distance between a pair of input fibers. Specifically, the input to the network is the FiberMaps of a fiber pair and a pseudo annotation of the fiber pair distance. For the pairwise fiber distance, we use the minimum average direct-flip (MDF) distance which is widely successful in WMFC \cite{C5,C7}. The computation of fiber distance considers the order of points along the fibers; thus, fiber distance is not affected if a fiber point sequence is flipped. A Siamese Network \cite{C23}, a neural network that encodes different inputs and computes comparable outputs with shared weights, is then adopted to learn embeddings of an input FiberMap pair and output Euclidean distance between the embeddings. The distance prediction loss $L_p$ is the mean squared error between embedding distance and fiber-distance pseudo annotations. By using fiber distances as pseudo annotations, the network is guided to generate similar embeddings for close fibers, even those with flipped point orders.

\subsection{Clustering Layer and Clustering Loss}
Here we adopt the DCEC model design \cite{C15}. In the clustering stage, a clustering layer is designed to encapsulate cluster centroids as its trainable weights and compute a soft assignment label $q_{ij}$ using Student’s t-distribution \cite{C27,C14}:
\begin{equation}
q_{ij} = (1+\left \| z_i-\mu_j \right \|^2)^{-1} / (\begin{matrix} \sum_{j'} (1+\left \| z_i-\mu_j' \right \|^2)^{-1} \end{matrix})
\end{equation}
where $\mathnormal{z}_i$ is the embedding of fiber $i$ and $\mu_j$ is the centroid of cluster $j$. $q_{ij}$ is the probability of assigning fiber $i$ to cluster $j$. The network is trained in a self-training manner and its clustering loss $L_c$ is defined as a KL divergence loss \cite{C14}:
$L_c = \mathit{KL}(P||Q)= \begin{matrix} \sum_{i} \begin{matrix} \sum_{j} p_{ij}log\frac{p_{ij}}{q_{ij}} \end{matrix} \end{matrix}$, where $ p_{ij}= ({q^2_{ij}/ \sum_{i} q_{ij})} / (\sum_{j'} (q^2_{ij'}/ \sum_{i} q_{ij'}))$. The distance prediction loss is retained in this stage, and the total loss is $L=L_p + \lambda L_c$, where $\lambda$ is the weight of $L_c$. During inference, a fiber $i$ is assigned to the cluster with the maximum $q_{ij}$.


\subsection{Incorporation of Anatomical Information and Outlier Removal}
We extend our proposed self-learning framework described above to enable two important tasks in WMFC, i.e., inclusion of additional anatomical information for anatomical coherence and removal or filtering of false positive outlier fibers. For the first task, we propose to incorporate Freesurfer parcellation \cite{C29} information during the clustering stage. We design a new soft label assignment probability definition which is used to calculate loss and extends Eq.(1) to further regularize that fibers within a cluster pass through the same brain regions:
\begin{equation}
q_{ij} = (1+\left \| z_i-\mu_j \right \|^2*(1-D_{ij}))^{-1} / ( \begin{matrix} \sum_{j'} (1+\left \| z_i-\mu_{j'} \right \|^2*(1-D_{ij'}))^{-1} \end{matrix} )
\end{equation}
where $D_{ij}$ is the Dice score between the set of Freesurfer regions of fiber $i$ and the set of Freesurfer regions of cluster $j$. We use the Tract Anatomical Profile (TAP) proposed in \cite{C7} to define the set of Freesurfer regions commonly intersected by the fibers in a cluster (at least 40\% of fibers, as in \cite{C7}). During training, the TAP is initially calculated from the clusters generated by $\mathnormal{k}$-means and is updated iteratively with new predictions during the training process. During inference, soft label assignments are calculated with Eq.(2) and fibers are assigned to the cluster with maximum $q_{ij}$, referred to as $q_m$.

For outlier removal, we remove fibers using the maximum label assignment probability $q_m$, considering that fibers with higher $q_m$ tend to have more confidence of belonging to the corresponding cluster and are less likely to be outliers. Therefore, we remove outliers by setting a threshold $h$ on the $q_m$ values of fibers, meaning that fibers with $q_m < h$ will be rejected from the final clusters.

\subsection{Implementation Details}
As shown in Fig.1, our model architecture includes three convolutional layers of sizes 5$\times$5$\times$32, 5$\times$5$\times$64 and 3$\times$3$\times$128, respectively, to extract feature maps. These feature maps are flattened to a vector, followed by a fully connected layer to compute embeddings with a dimension of 10 (suggested in \cite{C15}). In the pretraining and clustering stages, the network is trained for 25000 iterations with a learning rate of 0.0001 and another 4000 iterations with a learning rate of 0.00001, which are sufficient to achieve training convergence. Admax \cite{C28} is used for optimization in both stages. All experiments are performed on an NVIDIA RTX 2080Ti GPU using Pytorch (v1.7.1) \cite{pytorch}. The weight of clustering loss $\lambda$ is set to be 0.1, as suggested in \cite{C15}. We set the threshold $h$ for outlier removal to be 0.015 to reject fibers with extremely low cluster assignment probabilities.

\section{Experiments and Results}
\subsection{Dataset}
In our experiments, we used a dataset of 200 healthy adults from the Human Connectome Project \cite{C24}. 100 subjects were used for training, 50 for validation and 50 for testing. Tractography data were generated using a two-tensor unscented Kalman filter (UKF) method \cite{C30}, and tractography co-registration was performed using an affine followed by a nonrigid registration \cite{o2012unbiased}. Fibers longer than 40 mm were retained to avoid any bias towards implausible short fibers. For each training subject, 10,000 fibers were randomly selected, generating a training dataset of 1 million samples. For testing and validation, all whole-brain tractography fibers were used (around 500,000 per subject). Fibers were downsampled to 14 points \cite{C22} to obtain the FiberMap input to neural network. We performed diffusion MRI tractography and visualization in 3D Slicer (www.slicer.org) via the SlicerDMRI  project (http://dmri.slicer.org) \cite{norton2017slicerdmri,zhang2020slicerdmri}.

\subsection{Evaluation Metrics}
Three evaluation metrics were adopted to quantify performance of our proposed method and enable comparisons among approaches. The first one is the Davies–Bouldin (DB) index \cite{C25}, which is computed as:
\begin{equation}
\mathit{DB} = ({1}/{n}) \begin{matrix} \sum_{k=1}^n \mathop{max}_{x\neq y}(\frac{\alpha_i+\alpha_j}{d(c_i,c_j)}) \end{matrix}
\end{equation}
where $n$ is the number of clusters, $\alpha_i$ and $\alpha_j$ are mean pairwise intra-cluster fiber distances, and $d(c_i,c_j)$ is the inter-cluster fiber distance between centroids $c_i$ and $c_j$ of cluster $i$ and $j$ \cite{vazquez2020ffclust}. A smaller DB score indicates a better separation between clusters. The second metric is White Matter Parcellation Generalization (WMPG) \cite{C7}, which is used to represent the percentage of clusters successfully detected across the testing subjects. In our work, clusters with a over 10 fibers are considered to be successfully detected \cite{C7}. The last metric is Tract Anatomical Profile Coherence (TAPC) \cite{C7}, which measures if the fibers within a cluster c commonly pass through the same brain anatomical regions:

\begin{equation}
\mathit{TAPC}(c)= ( \begin{matrix} \sum_{f=1}^{\mathit{NF}(c)} Dice(\mathit{TAP}(f),\mathit{TAP}_\mathit{atlas}(c)) \end{matrix}) / {\mathit{NF}(c)}
\end{equation} 
Higher TAPC scores indicate better anatomical coherence.

\subsection{Evaluation Results}
\subsubsection{Comparison with State-of-the-art Methods} We compare our proposed approach with two open-source state-of-the-art WMFC algorithms, WhiteMatterAnalysis (WMA)\cite{C7} and QuickBundles (QB) \cite{C5}. WMA is an atlas-based WMFC method that shows high performance and strong correspondence across subjects. QB is a widely used WMFC method that performs clustering within each subject and achieves group correspondence with post-processing steps. We use the open-source software packages WMA v0.3.0 and Dipy v1.3.0 with their default settings. For all experiments, we perform WMFC into 800 clusters (which has been suggested to be a good whole brain tractography parcellation scale \cite{C7}). Dipy does not accept an input number of clusters; therefore, we tuned parameters in each subject to obtain a number as close as possible to 800 clusters (greater than or equal to 800). All results are reported using data from the 50 test subjects. The WMPG and TAPC metrics require corresponding clusters across all subjects; these are automatically generated by WMA and our proposed DFC method. For QB, correspondence is achieved by matching cluster centroids from all subjects to those of one selected subject (with exactly 800 clusters) according to the fiber distances between centroids, as suggested by the QB developers \cite{C5}. 


\begin{table}
\caption{Quantitative comparison results. SOTA: state of the art.}\label{tab1}
\centering
\begin{tabular}{l|c|l|l|l|l}
\hline
\quad & Methods &  DB index & WMPG & TAPC & Time(s)\\
\hline
\multirow{3}{*}{\tabincell{c}{SOTA\\Comparison}} & WMA & 3.231$\pm$0.153 & 99.22\%$\pm$0.79\% & 0.802$\pm$0.006 & 3210 \\
~ & QB &  \textbf{2.419$\pm$0.096} & 81.14\%$\pm$2.64\% &0.690$\pm$0.015 & 240 \\
~ & DFC & 2.661$\pm$0.107 & \textbf{99.35\%$\pm$0.54\%}&\textbf{0.836$\pm$0.006}&\textbf{205}\\
\hline
\multirow{2}{*}{\tabincell{c}{Baseline\\Comparison}}&DCEC &  15.661$\pm$4.390 & \textbf{99.87\%$\pm$0.35\%} &0.755$\pm$0.009&-\\
~&DFC & \textbf{2.661$\pm$0.107} & 99.35\%$\pm$0.54\%&\textbf{0.836$\pm$0.006}&-\\
\hline
\multirow{3}{*}{\tabincell{c}{~ Ablation\\\quad Study \qquad}}&DFC$_\mathit{no-fs-ro}$ &  3.095$\pm$0.156 & 99.80\%$\pm$0.48\% &0.773$\pm$0.009&-\\
~&DFC$_\mathit{no-ro}$ &  3.152$\pm$0.139 & \textbf{99.82\%$\pm$0.32\%} &0.816$\pm$0.007&-\\
~&DFC & \textbf{2.661$\pm$0.107} & 99.35\%$\pm$0.54\%&\textbf{0.836$\pm$0.006}&-\\
\hline
\end{tabular}
\end{table}

\begin{figure}[!t]
\includegraphics[width=0.85\textwidth]{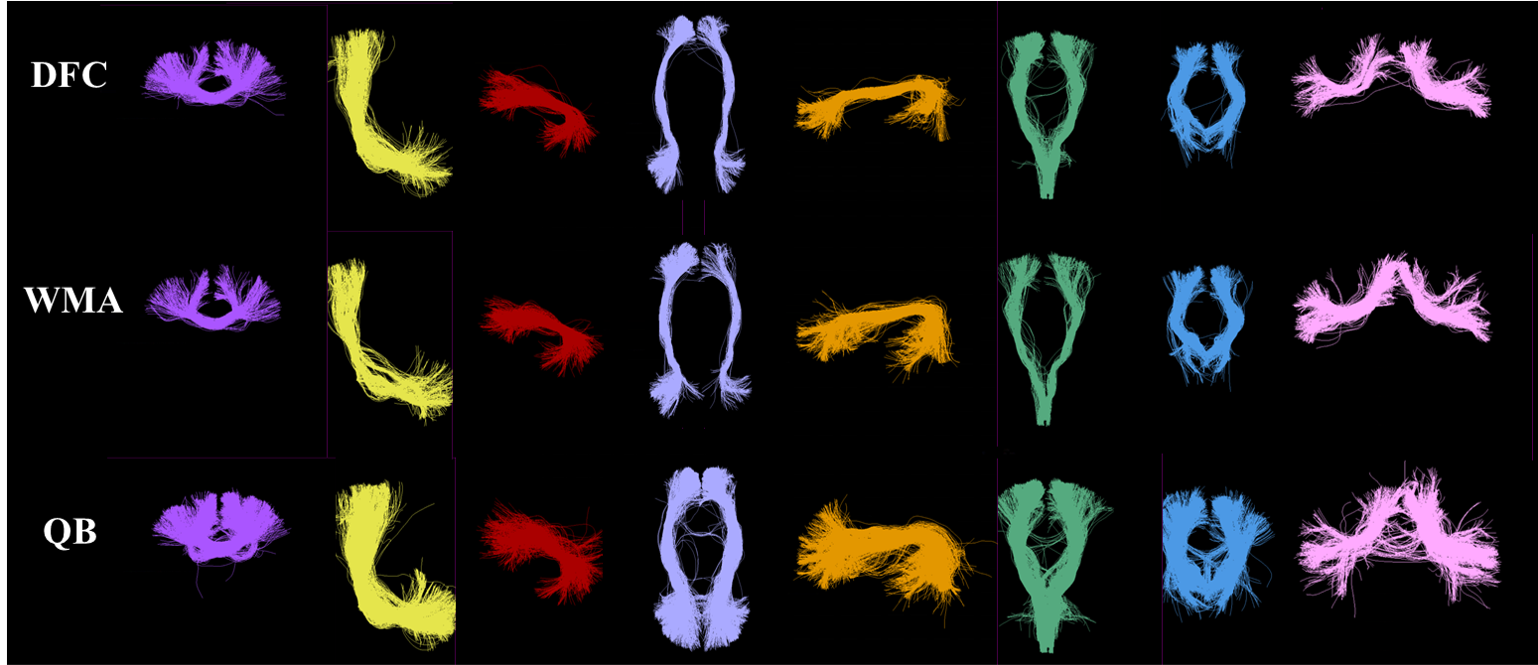}
\centering
\caption{Visualization of example clusters generated from DFC, WMA, and QB in one subject. Similar clusters were identified across methods for visualization.
} \label{fig2}
\end{figure}

As shown in Table \ref{tab1}, our DFC method exhibits the best performance in general. For the DB index metric, QB obtained a slightly lower value than DFC, likely because intra-cluster distances are lower when performing within-subject clustering since the obtained clusters do not describe anatomical variability across subjects. When compared to the atlas-based WMA, the DB index of our method is obviously smaller, indicating more compact and/or better separated clusters. As for WMPG, both our method and WMA successfully detected over 99$\%$ of clusters while the WMPG score of QB is around 80$\%$ indicating poor correspondence across subjects. The TAPC metric of DFC obtained the highest value among the three methods owing to the incorporation of anatomical information, indicating the best anatomical coherence of clusters. Fig. \ref{fig2} gives a visual illustration of obtained clusters for each method.

\begin{figure}[t]
\centering
\includegraphics[width=0.75\textwidth]{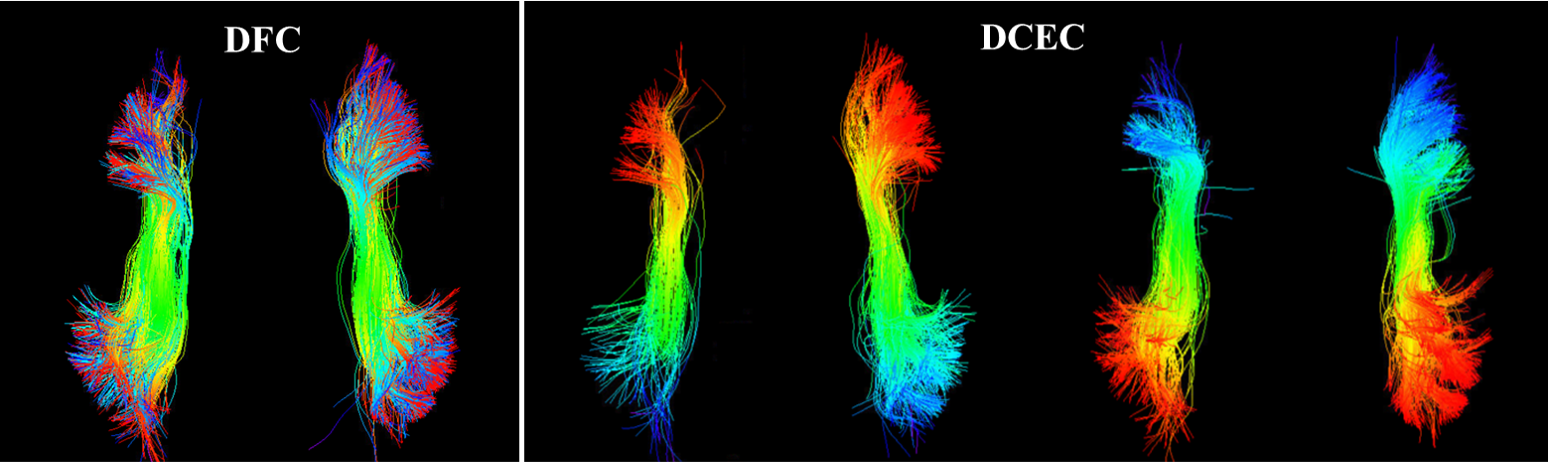}
\caption{Illustration of corresponding clusters from DFC and DCEC. Colors represent order of points along a fiber with red for starting point and blue for ending point.} \label{fig3}
\end{figure}

To evaluate efficiency of approaches, inference time of one subject is also recorded and shown in Table \ref{tab1}. All methods were tested on a computer equipped with a 2.1 GHz Intel Xeon E5 CPU (8 DIMMs; 32 GB Memory). For fair comparison, DFC was set to run on CPU instead of GPU. The results show that our method is much faster than WMA and slightly better than QB.
\subsubsection{Comparison with the Baseline Method} We also compared our proposed method with the DCEC baseline model. The results in Table \ref{tab1} show a large improvement of the DB index of our method compared to DCEC, because DCEC separately clusters fibers with close positions but flipped point orders. As shown in Fig. \ref{fig3}, spatially close fibers with different point orders are split into two clusters in DCEC, while our proposed DFC method groups them together.

\subsubsection{Ablation Study} 
\begin{figure}[t]
\centering
\includegraphics[width=0.75\textwidth]{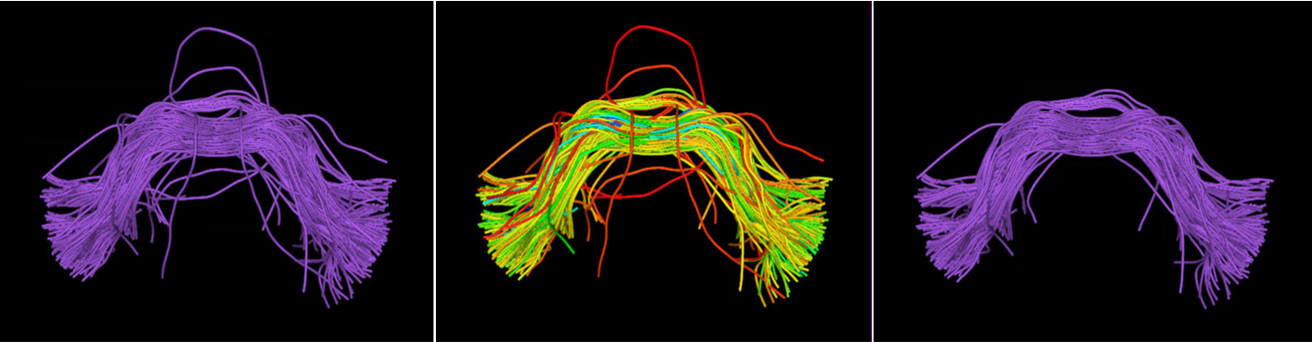}
\caption{Illustration of outlier removal process. Left: cluster before outlier removal; Middle: fiber soft label assignment probability (rainbow coloring with red representing 0); Right: cluster after outlier removal.} \label{fig4}
\end{figure}

We performed an ablation study to investigate how different factors influence performance of our method. Evaluation of three models was performed, including  DFC$_{no-fs-ro}$ (DFC without FreeSurfer information and outlier removal), DFC$_{no-ro}$ (DFC without outlier removal but with FreeSurfer information) and DFC$_{proposed}$, as shown in Table \ref{tab1}. By adding FreeSurfer information into the model, the DB index and WMPG metrics do not show much difference, while the TAPC score exhibits obvious improvement. With implementation of outlier removal, the DB index and TAPC improve obviously, while WMPG shows slight decrease, which is inevitable due to the decreased number of fibers (but it still remains a high percentage). These results demonstrate effectiveness of our designed modules. As shown in Fig. \ref{fig4}, outlier fibers have apparently low values of soft label assignment probabilities and are then removed.

\section{Conclusion}
In this paper, we present a novel unsupervised deep learning framework for dMRI tractography WMFC. We adopt the self-supervised learning strategy to enable joint deep embedding and cluster assignment. Our method can handle several key challenges in WMFC methods, including handling flipped order of points along fibers, incorporating anatomical brain segmentation information, false positive fiber filtering and inter-subject correspondence of fiber clusters. Our results show advantages over clustering performance as well as efficiency compared to the state-of-art algorithms. Further research could be conducted to improve the framework, such as designing more complex network architectures, incorporating additional sources of anatomical information and balancing anatomical and fiber geometry information for clustering.

\clearpage

%
%
%
\bibliographystyle{splncs04}
\bibliography{mybibliography}
\end{document}